\documentclass[10pt,twocolumn,letterpaper]{article}

\usepackage{cvpr}
\usepackage{times}
\usepackage{epsfig}
\usepackage{graphicx}
\usepackage{amsmath}
\usepackage{amssymb}
\usepackage{booktabs}

\usepackage{algorithm,algpseudocode}
\usepackage{amsfonts}

\usepackage{url}

\usepackage{subcaption}
\usepackage{graphicx}

\usepackage[pagebackref=true,breaklinks=true,letterpaper=true,colorlinks,bookmarks=false]{hyperref}

\cvprfinalcopy 

\def\cvprPaperID{10630} 

\ifcvprfinal\fi
\begin{document}

\title{Efficient Micro-Structured Weight Unification and Pruning for Neural Network Compression}

\author{Sheng Lin\textsuperscript{\rm 1}, 
Wei Jiang\textsuperscript{\rm 1}, 
Wei Wang\textsuperscript{\rm 1}, 
Kaidi Xu\textsuperscript{\rm 1,2}, 
Yanzhi Wang\textsuperscript{\rm 2}, 
Shan Liu\textsuperscript{\rm 1}, 
Songnan Li\textsuperscript{\rm 1}\\
\textsuperscript{\rm 1}Tencent Media Lab, 
\textsuperscript{\rm 2}Northeastern University, \\
{\tt\small barneylin@tencent.com, vwjiang@tencent.com rickweiwang@tencent}
}

\maketitle


\begin{abstract}

Compressing Deep Neural Network (DNN) models to alleviate the storage and computation requirements is essential for practical applications, especially for resource limited devices. Although capable of reducing a reasonable amount of model parameters, previous unstructured or structured weight pruning methods can hardly truly accelerate inference, either due to the poor hardware compatibility of the unstructured sparsity or due to the low sparse rate of the structurally pruned network. Aiming at reducing both storage and computation, as well as preserving the original task performance, we propose a generalized weight unification framework at a hardware compatible micro-structured level to achieve high amount of compression and acceleration. Weight coefficients of a selected micro-structured block are unified to reduce the storage and computation of the block without changing the neuron connections, which turns to a micro-structured pruning special case when all unified coefficients are set to zero, where neuron connections (hence storage and computation) are completely removed. In addition, we developed an effective training framework based on the alternating direction method of multipliers (ADMM), which converts our complex constrained optimization into separately solvable subproblems. Through iteratively optimizing the subproblems, the desired micro-structure can be ensured with high compression ratio and low performance degradation. We extensively evaluated our method using a variety of benchmark models and datasets for different applications. Experimental results demonstrate state-of-the-art performance.
\end{abstract}

\section{Introduction}
Deep Neural Networks (DNNs) have shown great success in solving a wide range of tasks in computer vision, audio recognition, etc. However, practical deployment of large DNNs on resource constrained devices remains a major challenge. Active research has been conducted recently on DNN model compression and acceleration \cite{courbariaux2015binaryconnect,han2015learning,hinton2015distilling}. Among these methods, weight pruning ~\cite{han2015learning}, weight quantization~\cite{courbariaux2015binaryconnect} and knowledge distillation~\cite{hinton2015distilling} are most popular, with proven ability to reduce the model size and maintain the original prediction performance. These techniques are of particular interests to both software and hardware optimization of DNN inference engines. 

{\em Weight pruning} utilizes the fact that a large portion of weights in a trained DNN are close to zero. A representative retraining process is used to purposely increase the sparsity of weights. The pioneer work~\cite{han2015learning} proposed a heuristic and iterative weight pruning method, which reduces weight parameter and maintains a decent accuracy. There are many extended works~\cite{dai2017nest,dong2017learning,guo2016dynamic,yang2016designing} with more advanced algorithms to achieve better compression rate. However, such irregular or non-structured weights can hardly be used for inference acceleration in hardware devices, since the weight matrices after irregular pruning are not compatible with current parallel execution hardware designs. This shortcoming has been confirmed by the throughput drop in recent works~\cite{wen2016learning, li2016pruning}, which proposed to integrate the regularity in weight pruning to achieve more structured weight representation. The structured approaches, such as  filter  pruning,  channel  pruning,  and  filter  shape  pruning are hardware friendly by inducing sparsity pattern to targeted inference engine. The weight matrices after structured pruning usually can be compacted to dense matrices, and therefore are compatible with off-the-shelf libraries on CPUs and GPUs.

\emph{Weight quantization} leverages the redundancy in the number of bits to represent weights in DNNs~\cite{courbariaux2015binaryconnect}. Weight quantization is hardware friendly, and is well supported in FPGA, ASIC, GPU and mobile devices. Both storage and computation of DNNs can be reduced proportionally to the number of bits used in these hardware platforms. Moreover, for extremely low bit design such as binary and ternary weight quantization, bit shift operations can substitute multiplication operations which can extremely accelerate  inference. However, the low bits weight representation usually hurts the model performance significantly, especially for deep and large DNN models.  




Based on the pros and cons of different model compression methods, in this paper, we propose a new micro-structured weight unification method to reduce the model storage and computation, and maintain the original task performance. The main idea  is to reduce the weight representation in a structured way that can benefit the weight storage and the general matrix multiplication (GEMM) computation. Our micro-structured weight unification approach unifies weights within a selected weight block, where all weight coefficients in that selected block will share the same absolute value. Compared with weight pruning methods, we maintain the neuron connections of unified weights (being set to a unified value) instead of removing them (being set to zero), to better maintain the original network structure and preserve the model performance. Our micro-structured weight unification is hardware friendly for both model storage and inference computation, in the sense of accommodating flexible micro-structured block shapes that are compatible with the underlying inference engine. 

From another perspective, when the unified value is zero, we get a micro-structured weight pruning approach as a special case of our micro-structured weight unification method. The micro-structured weight unification provides a more balanced model for both model compression and task performance, while micro-structured weight pruning pursues more aggressive compression effects. They have different strengths on different models for different tasks, and can be combined to further improve the overall performance.

In addition, targeting at the non-differentiable optimization caused by the hard constraints in micro-structured weight unification and/or micro-structured weight pruning, we exploit the alternating direction method of multipliers (ADMM) to ensure effective training. ADMM has been successfully used for weight pruning to achieve large compression ratios recently~\cite{leng2017extremely, zhang2018structadmm, zhang2018systematic}. Inspired by these works, we propose a systematic framework for micro-structured weight unification and micro-structured weight pruning using ADMM. Our framework ensures the feasibility of a quality solution with high compression rate and decent task performance, and our method consistently performs well on different models for various tasks.


We conducted extensive experiments to evaluate our method on several DNN models over several datasets, such as CIFAR-100, ImageNet and DCASE. Following the convention of prior arts in model compression, our experiments focus on the convolutional layers and fully-connected layers, which are the most common yet most computation intensive layers in most DNNs (e.g., ResNet, MobileNet). We note that our micro-structured weight unification and micro-structured weight pruning framework is model-agnostic and can be generally applied to other types of layers.

In summary, our main contribution is three fold:

\begin{itemize}

\item We propose a micro-structured weight unification method with flexible micro-structured block shapes. The total compression ratio and computation reduction are controlled by the block shape and unification ratio. 

\item We formulate the micro-structured weight unification as a general constrained weight optimization problem, and show that micro-structured weight pruning is a special case of this problem. These methods (micro-structured weight unification and pruning) are complementary to each other and can be potentially combined to further improve the overall performance.

\item We develop an effective training process using the ADMM optimization algorithm, which gives a high-quality feasible solution to our hard constrained micro-structured weight unification or micro-structured weight pruning problem. Our training framework can be generally applied to various models and tasks. 



\end{itemize}

\section{Related Work}



\subsection{Weight Pruning}
Weight pruning aims to reduce redundant or less important weights in DNNs. There are typically two types of weight pruning, non-structured weight pruning~\cite{han2015learning} and structured weight pruning~\cite{wen2016learning}.


\textbf{\emph{Non-structured weight pruning.}}  This weight pruning method is straightforward for model compression, since it prunes the weight value to zero and the removed weights will not be stored or involved in computation. The prior work of~\cite{han2015learning} reduced the number of parameters by 9$\times$ in AlexNet and 13$\times$ in VGGNet-16. Note that most of weight reduction in these networks is achieved in fully connected layer. Extensive later works, such as~\cite{dai2017nest, frankle2018lottery, zhang2018systematic}, further optimized the algorithm and improved the compression ratio. However, such non-structured pruning methods introduce a lot of overhead in both weight storage and matrix computation due to its irregularity in weight structure.


\textbf{\emph{Structured weight pruning.}} To overcome the limitation of non-structured weight pruning, SSL~\cite{wen2016learning} proposed to regularize weight structure at the levels of filters, channels, filter shapes, and layer depths.  More recently, ~\cite{he2017channel} proposed an iterative two-step algorithm by an $L_{1}$ regularization to remove less important filters and reconstruct the outputs with remaining channels with linear least squares. These methods are effective in convolutional layers, but cause large performance drop on fully-connected layers, since removing multiple rows or columns can cause significant information loss. Therefore, the overall compression ratio (with acceptable performance drop) after structured pruning is much lower than non-structured methods.

\textbf{\emph{Fine-grained structured weight pruning.}} Several fine-grained structured pruning methods were investigated recently~\cite{jang2019simd, niu2020patdnn, van2019rethinking}, where more flexible structures were integrated to regularize weight sparsity. 
The block pruning~\cite{van2019rethinking} studied how to group neighboring weights at the scale of blocks, but it only focused on fully connected layers rather than convolutional layers.
~\cite{jang2019simd} designed an SIMD-aware pruning method which divided weight matrix by tailored block size for SIMD support, and each block was pruned independently. This hybrid sparsity can fully leverage the multi-level parallel hierarchy in modern CPU. However, it can only be applied to devices that support SIMD operation.
The latest NVIDIA A100 supports a 2:4 structured sparsity 
to double the throughput for its Sparse Tensor Core~\cite{nvidai_a100}.  
PatDNN~\cite{niu2020patdnn} proposed kernel pruning where connectivity among kernels was structurally pruned based on different masked patterns. 
The fine-grained sparsity allows a better trained model with a better compression ratio.



\subsection{Weight Quantization}
Using fewer bits to represent weights in DNNs, weight quantization takes advantage of the inherent redundancy in weight representation to reduce model size, and therefore directly accelerates inference speed. Weight quantization methods can be roughly divided into two categories: linear and non-linear quantization schemes, featuring uniformly spaced or unevenly spaced quantization levels, respectively.

Many previous works focused on linear quantization of weights to binary, ternary, or fixed-point. Binary and ternary quantization utilize low-bit weight representation to improve inference computing efficiency by eliminating multiplications. The family of binary and ternary quantization schemes were implemented in BNN~\cite{courbariaux2016binarized}, XNOR-Net~\cite{rastegari2016xnor}, TWN~\cite{li2016ternary}, and DoReFa-Net~\cite{zhou2016dorefa}. However, they suffer from substantial performance loss, especially for complicated tasks or network structures. To maintain better performance, fixed-point quantization uses more bits to represent weights,  such as PACT~\cite{choi2018pact} with an added activation clipping parameter, DSQ~\cite{gong2019differentiable} with a series of hyperbolic tangent functions, and  QIL~\cite{jung2019learning} with parameterized interval. The quantization levels are unequal in non-linear schemes which are mostly logarithmic. The work of \cite{miyashita2016convolutional} first proposed to encode weights in the log-domain. Hence, the multiplication of input and weight matrix can be easily replaced by bit shift operations.

\subsection{ADMM for Weight Pruning/Quantization}

ADMM is a powerful optimization algorithm, which decomposes an original complex problem into subproblems that can be solved separately and iteratively~\cite{boyd2011distributed}. It is usually used to accelerate the convergence of convex optimization problems and enable distributed optimization. As a proven property, ADMM can effectively deal with the problem that subjects to a subset of combinatorial constraints. For example, suppose we want to solve the optimization problem $\min_{\bf{x}} f({\bf{x}})+g({\bf{x}}).$ This problem can be solved as a series of unconstrained minimization problems by ADMM, by which the problem is further decomposed into two subproblems on $\bf{x}$ and $\bf{z}$ (auxiliary variable). The first subproblem derives $\bf{x}$ given $\bf{z}$: $\min_{\bf{x}} f({\bf{x}})+q_1(\bf{x}|\bf{z})$. The second subproblem derives $\bf{z}$ given $\bf{x}$: $\min_{\bf{z}} g({\bf{z}})+q_2(\bf{z}|\bf{x})$. Both problems will be solved iteratively until convergence.

Recent work~\cite{leng2017extremely, zhang2018structadmm, zhang2018systematic} have incorporated ADMM for DNN weight pruning and weight quantization. ADMM training was employed in ~\cite{leng2017extremely} to increase the accuracy of extremely low bit width DNNs (binary network and ternary network). Zhang~\emph{et al.}~\cite{zhang2018systematic} formulated weight pruning as a constrained optimization problem, which employed the cardinality function to induce  non-structured weight pruning. Although a high compression ratio was achieved, this method suffered from the non-structured irregularity. The framework was further extended to structured weight pruning~\cite{zhang2018structadmm}, yet with limited compression ratio.


\section{Our Method}
\label{sec:method}

We first formulate our problem. The loss function associated with an DNN can be denoted by $f \big( \{{\bf{W}}_{i}\}_{i=1}^N \big)$, which will be minimized in the training process:
\begin{equation}
\small
\label{eq:problem}
\begin{aligned}
& \textit{L}_p  = \underset{ \{{\bf{W}}_{i} \}}{\text{min}}
& & f \big( \{{\bf{W}}_{i} \}_{i=1}^N \big),\\ 
& \text{subject to}
& & {\bf{W}}_{i}\in {\bf{\mathcal{S}}}_{i}, \; i = 1, \ldots, N, \\
\end{aligned}
\end{equation}

\noindent where ${\bf{W}}_{i}$ is the weight tensor of the $i$-th layer, and  ${\bf{\mathcal{S}}}_{i}$ is the set of all possible solutions of ${\bf{W}}_{i}$ satisfying some target constraints. For example, the unstructured pruning methods use the $l0$ or $l1$ norm of ${\bf{W}}_{i}$ as a soft regularization to promote weight sparsity, and the structured pruning methods enforces constraints of all-zero channels/filters of ${\bf{W}}_{i}$.

\subsection{Micro-Structured Weight Unification}
\label{sec:swu}

In this section we propose a micro-structured weight unification constraint. Our goal is to train the model so that weight coefficients in each of the selected micro-structured blocks share the same absolute value. The block shape is selected to be compatible with the underlying GEMM computation so that the unified micro-structured blocks can reduce weight storage and computation.

Specifically, we divide each ${\bf{W}}_{i}$ into $M$ blocks based on a block shape $\bf{P}$, such as ${D_1}\times{D_2}\times{D_3}$ 3-dim blocks or ${D_1}\times{D_2}$ 2-dim blocks (or 1-dim blocks if ${D_1}$ or ${D_2}$ equals to 1). If selected to be unified, all coefficients in the $j$-th selected block of ${\bf{W}}_{i}$ will be set to have a unified absolute value $q_{i,j}$, while keeping their original signs:
\begin{equation}
\label{eq:value}
     v_{i,j,l} = \begin{cases}
     q_{i,j} & \text{if $w_{i,j,l}$ $\geq$ 0,}\\
     -q_{i,j} & \text{otherwise.}
     \end{cases}  
\end{equation}
$v_{i,j,l}$ is the newly assigned value to the original $l$-th weight parameter $w_{i,j,l}$ in the $j$-th block of ${\bf{W}}_{i}$. The unified absolute value $q_{i,j}$ can be computed as the average of the absolute value of all weights in this $j$-th block. Since the block shape aligns with the underlying GEMM computation, we can reduce the model storage and inference computation of the unified blocks. 

In a special situation, we can choose $q_{i,j}\!=\!0$ and set all weight coefficients $v_{i,j,l}$ in the selected block to be zero. This results in the micro-structured weight pruning approach, which is an extreme special case of the micro-structured weight unification, where we can skip the storage and inference computation of the pruned blocks completely. 

\begin{figure}
\centering
\begin{subfigure}[b]{0.38\textwidth}
   \includegraphics[width=1\linewidth]{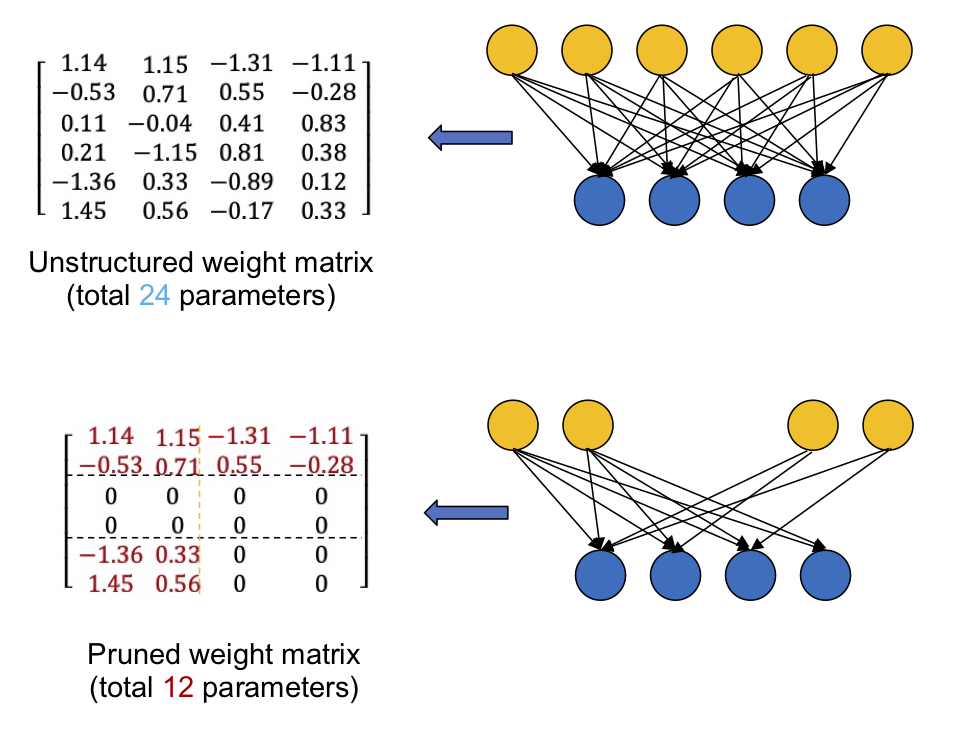}\vspace{-.5em}
   \caption{Micro-structured weight pruning}
   \label{fig:2x2_pruning} 
\end{subfigure}

\begin{subfigure}[b]{0.5\textwidth}
   \includegraphics[width=1\linewidth]{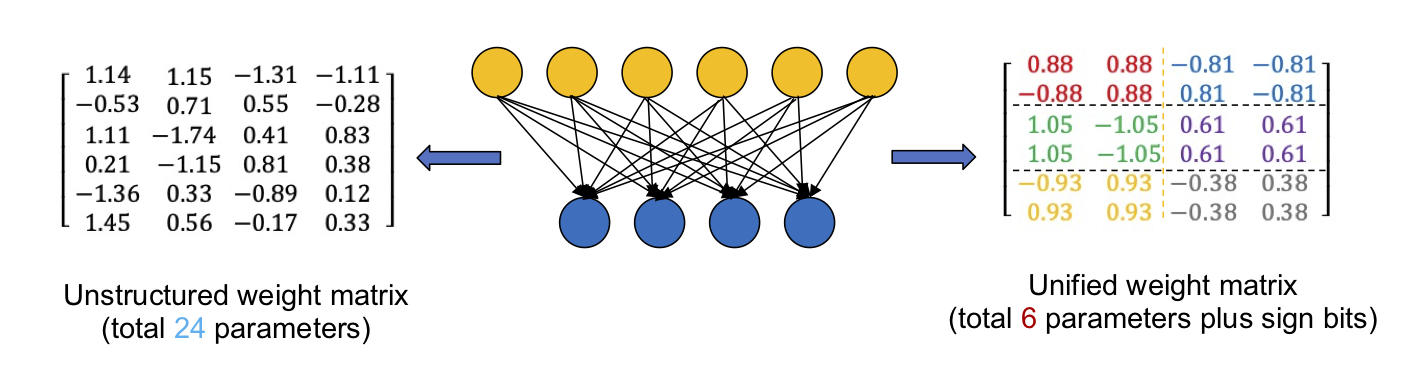}\vspace{-.5em}
   \caption{Micro-structured weight unification}
   \label{fig:2x2_unification}
\end{subfigure}

\caption{Example of model reduction with micro-structured unification and micro-structured pruning}\vspace{-.5em}
\end{figure}

Figure~\ref{fig:2x2_pruning} and Figure~\ref{fig:2x2_unification} give an example illustration of the micro-structured weight pruning and micro-structured weight unification, respectively. Given a fully-connected layer with ${6}$ input and ${4}$ output neurons, we divide the weight matrix into ${2}\times{2}$ blocks. For micro-structured weight pruning in Figure~\ref{fig:2x2_pruning}, we prune those blocks having smallest $L_1$ norms of their weights. With a $50\%$ pruning ratio, we can get ${2}\times$ parameter reduction. For micro-structured weight unification in  Figure~\ref{fig:2x2_unification}, we unify all blocks (with $100\%$ unification ratio) in this weight matrix, and we can achieve nearly ${4}\times$ parameter reduction. 

From this example we can clearly see the pros and cons of these two methods. micro-structured weight unification keeps the original connections between neurons, which helps to maintain the original model capacity and therefore preserve the original task performance. As a result, a large unification ratio can be achieved without large performance degradation. In comparison, micro-structured weight pruning pursues model reduction more aggressively by removing neuron connections in the pruned blocks. This usually leads to large impact on the original task performance, and therefore limits the tolerable pruning ratio we can reach.

\begin{figure*} [t]
    \centering
     \includegraphics[width=2.0\columnwidth]{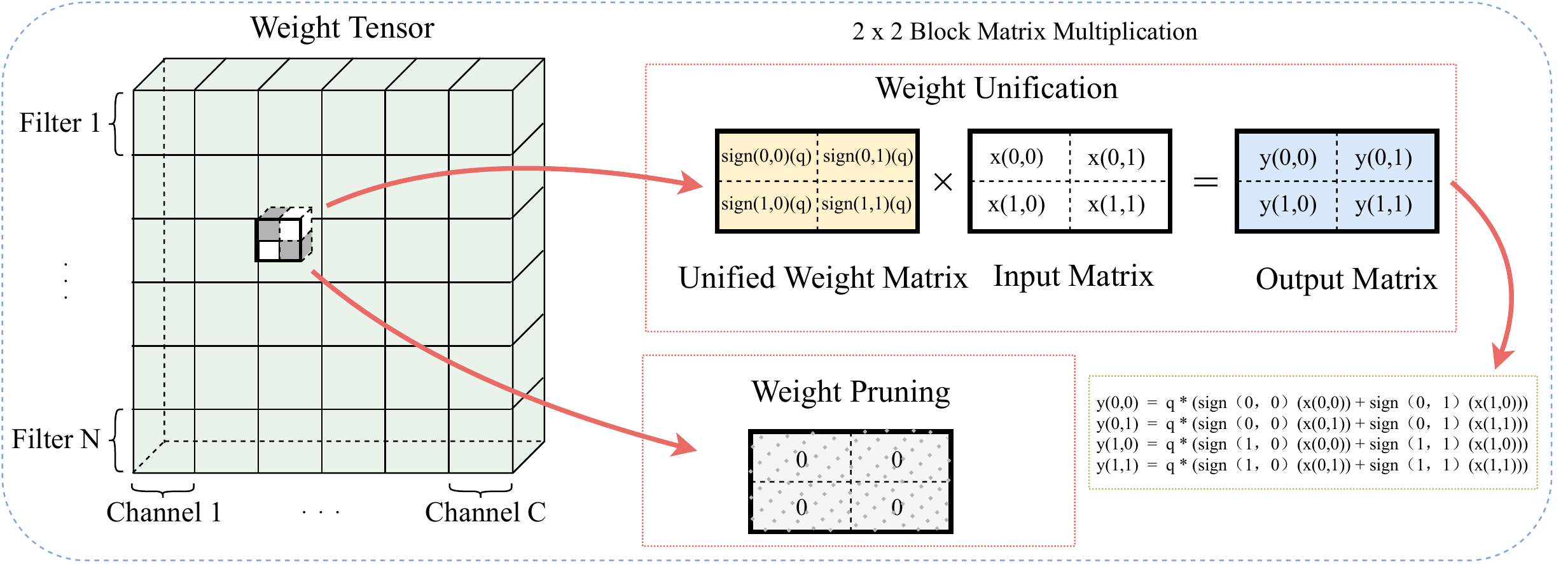} 
     \caption{Illustration of micro-structured weight unification and micro-structured weight pruning acceleration}
    \label{fig:multiplication}
\end{figure*}



\subsection{GEMM Compatible Block Shapes}

One most prominent feature of our micro-structured weight unification and micro-structured weight pruning is the flexibility to incorporate different block shapes, especially those compatible with parallel computing in hardware platforms such as CPU, GPU or FPGA. We can use block shapes that align with the design of hardware computing unit and memory storage and at the same time, are suitable for parallel computing to achieve inference acceleration.

Figure~\ref{fig:multiplication} gives an example of how micro-structured weight unification and micro-structured weight pruning can be used for accelerating computation. The inference convolution of a weight tensor is usually implemented by GEMM, and the GEMM computation is performed by dividing the matrices into small blocks. For micro-structured weight pruning, the GEMM acceleration is straightforward where pruned all-zero blocks can be completely skipped from the computation. For micro-structured weight unification, the GEMM acceleration is achieved by reducing the number of multiplication operations in the unified blocks. This is done by sharing temporary results in computation. Taking a $2 \times 2$ block shape as an example, as shown in the figure, the original matrix multiplication process needs $8$ multipliers, while the unified block only needs $4$ multipliers.

Different block shapes usually have different amounts of storage and computation reduction. Table~\ref{tab:patternshape} lists the corresponding information of a few block shapes used in our experiments, and their detailed performance will be discussed in later Section~\ref{experiment:blockshape}. Generally speaking, we can choose 2-dim blocks or 3-dim blocks. With the same compression ratio (the same amount of parameter reduction), 2-dim blocks usually give better computation acceleration. With the same task performance, 3-dim blocks can be more flexible, leading to better compression ratio. Also, the larger the blocks, the more reduction we can get, but the more strict constraints we put on the weights, and therefore the larger performance drop we have for the original task. 

Comparing micro-structured weight unification and weight pruning, with the same number of blocks being processed (unified or pruned), micro-structured weight pruning gives more parameter reduction. However, as discussed before, micro-structured weight unification better preserves the neuron connections and hence better maintains the original task performance. The optimal method (unification or pruning) and the optimal block shape may vary based on different model architectures for different tasks.

\begin{table}[thb]
\caption{Example for different block shapes}
\centering
\setlength{\abovecaptionskip}{0pt}%
\setlength{\belowcaptionskip}{8pt}%
\label{tab:patternshape}
\resizebox{\columnwidth}{!}{%
\begin{tabular}{c|cc}
\toprule
Block Shape & Multiplier Reduction & Storage Reduction \\ 
\midrule
{[}2, 2{]}    & {$\times$} 2                     & {$\times$} 4                 \\
{[}4, 1{]}    & {$\times$} 4                     & {$\times$} 4                 \\
{[}8, 1{]}    & {$\times$} 8                     & {$\times$} 8                 \\
{[}2, 2, 2{]} & {$\times$} 2                     & {$\times$} 8                \\
\bottomrule
\end{tabular}
}
\vspace{-5mm}
\end{table}

\subsection{Training with ADMM}
\label{sec:admm}



The problem (\ref{eq:problem}) enforces the weight unification or weight pruning block shapes by hard constraints, which makes it difficult to optimize directly. In this subsection, we use the ADMM algorithm to effectively address the composite constraints and optimize an alternative auxiliary problem in an iterative fashion.


Specifically, the original problem (\ref{eq:problem}) can be equivalently rewritten as:
\begin{equation}
\small
\label{admm_form}
\begin{aligned}
& \textit{L}_p  =  \underset{ \{{\bf{W}}_{i}\}}{\text{min}}
& & f \big( \{{\bf{W}}_{i} \}_{i=1}^N \big)+\sum_{i=1}^{N} h_{i}({\bf{Q}}_{i}),
\\ & \text{subject to}
& & {\bf{W}}_{i} = {\bf{Q}}_i, \; i = 1, \ldots, N.
\end{aligned}
\end{equation}

\noindent The notation $h_{i}(\cdot)$ is the indicator function, where $h_{i}(\bf{Q}_i)=0$ if $\bf{Q}_i$ meets the constraint, otherwise $h_{i}(\bf{Q}_i)=+\infty$.

\begin{algorithm}[t] 
\caption{The Process of ADMM training}
\label{alg: admm}
\begin{algorithmic}[1]
\Require{Pre-trained Weight matrix ${\bf{W}}_0$, micro-structured weight unification function $\bf{Unify()}$}
\Ensure{Unified weight matrix ${\bf{Q}}_{K}$}

\Statex

\State{Initialize ${\bf{Q}}_{0}$ = ${{\bf{Unify}}({\bf{W}}_0)}$}
\State{Initialize ${\bf{U}}_0$ to zero tensor}
\For{$k = 1$ to $K$}
    \State{Calculate the regularization term by ${\bf{Q}}_{k}$ and ${\bf{U}}_{k}$}
    \State{Optimize training loss by SGD or Adam and get updated ${\bf{W}}_{k+1}$}
    \State{${\bf{Q}}_k$ = ${\bf{Unify}}({\bf{W}}_{k+1}+{\bf{U}}_{k})$ }
    \State{${\bf{U}}_{k+1}$ = ${\bf{U}}_{k}$ + ${\bf{W}}_{k+1}$ - ${\bf{Q}}_{k+1}$}
    \State{Adjust penalty factors $\rho_{i}$ in each iteration}
\EndFor

\end{algorithmic}
\end{algorithm}

By introducing the auxiliary variables ${\bf{Q}}_{i}$, dual variables ${\bf{U}}_{i}$ and penalty factors $\rho_{i}$, we can apply ADMM to further decompose problem (\ref{admm_form}) into simpler subproblems. The augmented Lagrangian formation of problem (\ref{admm_form}) is:
\vspace{-0.50em}
\begin{equation}
\small
\begin{aligned}
\label{equ7}
\textit{L}_p  =  \underset{\{{\bf{W}}_{i} \}}{\text{min}}
\ \ \ & f \big( \{{\bf{W}}_{i} \}_{i=1}^N \big) +  \sum_{i=1}^{N} \frac{\rho_{i}}{2}  \| {\bf{W}}_{i}-{\bf{Q}}_{i}+{\bf{U}}_{i} \|_{F}^{2}. \\
\end{aligned}
\end{equation}
The first term in problem (\ref{equ7}) is the differentiable loss function of the DNN and the second term is a quadratic regularization term of ${\bf{W}}_{i}$, which is differentiable and convex.
 
The ADMM algorithm~\cite{boyd2011distributed} iteratively optimizes each of the subproblems alternately. At iteration $k$, we carry out three steps in ADMM. In the first step,  we fix ${\bf{Q}}_i$ and ${\bf{U}}_i$ to minimize $\textit{L}_p$ over ${\bf{W}}_i$ as follows:
\vspace{-0.50em}
\begin{equation}
\small
    {\bf{W}}_{i}^{k+1} := \underset{ {\bf{W}}_{i}}{\text{arg min}}\quad \textit{L}_p(\{{\bf{W}}_i\}, \{{\bf{Q}}_i^k\}, \{{\bf{U}}_i^k\}),
\label{itera1}
\end{equation}
The subproblem (\ref{itera1}) can be solved by gradient-based optimization algorithms, such as the SGD or ADAM optimizer.

In the second step,  we fix ${\bf{W}}_i$ and ${\bf{U}}_i$ to minimize $\textit{L}_p$ over ${\bf{Q}}_i$ as follows:
\vspace{-0.50em}
\begin{equation}
\small
    {\bf{Q}}_{i}^{k+1} := \underset{ {\bf{Q}}_{i}}{\text{arg min}}\ \textit{L}_p(\{{\bf{W}}_i^{k+1}\}, \{{\bf{Q}}_i\}, \{{\bf{U}}_i^k\}),
\label{itera2}
\end{equation}
The subproblem (\ref{itera2}) is a convex quadratic problem and can be solved by Euclidean projection.

Finally, the dual variables $\bf{U}_i$ are accumulated and updated as follows:
\vspace{-0.50em}
\begin{equation}
\small
    {\bf{U}}_{i}^{k+1} := {\bf{U}}_{i}^{k}+{\bf{W}}_{i}^{k+1}-{\bf{Q}}_{i}^{k+1}.
\label{itera3}
\end{equation}

These subproblems are solved iteratively until reaching the convergence. The pseudo-code for the ADMM training process can be found in Algorithm~\ref{alg: admm}.

\section{Experiments}

We use Pytorch~\cite{paszke2017pytorch} for experiments. All training were performed using an NVIDIA DGX Station with 4 Tesla V100 GPUs (24 GB memory each). We evaluate our methods on several different datasets for different tasks.

\subsection{Datasets}


We used CIFAR-100, ImageNet, and DCASE for experiments.  CIFAR and ImageNet are classical datasets for computer vision, and DCASE is an audio classification dataset.  
 
\textbf{\emph{CIFAR-100:}} The CIFAR-100 dataset consists of tiny RGB images. It has 100 classes and contains 50K training images and 10K test images~\cite{krizhevsky2009learning}.

\textbf{\emph{ImageNet:}} The ImageNet ILSVRC2012 dataset has 1000 classes. A portion of the training set (comprising of 50 randomly sampled images from each class) is used for  validation. The original  RGB images are resized to $256\!\times\!256$ size from which a $224\!\times\!224$ patch is cropped as the input for DNNs.The reported accuracy of ImageNet is measured on the original validation set~\cite{imagenet_cvpr09}.


\textbf{\emph{DCASE:}} The DCASE2017 Task1 dataset consists of recordings from various acoustic scenes characterizing the acoustic environment. It is used for acoustic scene classification where the test recording is classified to one of the 15 predefined acoustic scene classes, such as park, pedestrian street, metro station and so on~\cite{mesaros2019sound}.

\subsection{Models}
We evaluated our micro-structured weight unification and micro-structured weight pruning approaches on VGGNet-16~\cite{simonyan2014very}, ResNet-50~\cite{he2016deep}, and MobileNet-V2~\cite{sandler2018mobilenetv2} using the ImageNet dataset for image classification, on Autoencoder using the CIFAR-100 dataset for image compression, and on ConvNet using the DCASE dataset for audio scene classification.


The evaluated VGGNet-16, ResNet-50, and MobileNet-V2 are standard pre-trained versions as in the corresponding references. The Autoencoder has 4 convolutional encoder layers and 5 convolutional decoder layers. The architecture of the Autoencoder is: $64
\textit{Conv}3\!\times\!3-32\textit{Conv}3\!\times\!3-32\textit{Conv}3\!\times\!3-16\textit{Conv}3\!\times\!3- 64\textit{Conv}3\!\times\!3-32\textit{Conv}3\!\times\!3-32\textit{Conv}3\!\times\!3-16\textit{Conv}3\!\times\!3-3\textit{Conv}3\!\times\!3$, where \textit{Conv} stands for convolutional layer, the number before \textit{Conv} (64, 32, etc.) is the filter number, and $3\!\times\!3$ is the kernel size. The ConvNet is a simple convolutional neural network with 2 convolutional layers, 2 batch normalization layers and 2 fully-connected layers, and the architecture can be summarized as: $32\textit{Conv}7\!\times\!7-\textit{BN}-64\textit{Conv}7\!\times\!7-\textit{BN}-\textit{FC}128\!\times\!100-\textit{FC}100\!\times\!15$, where \textit{BN} and \textit{FC} stand for batch normalization and fully-connected layers, respectively.

These ImageNet, CIFAR-100 and Dcase datasets and corresponding tested models are believed to be representative for evaluating model compression techniques, and are chosen by the MPEG standardization group to develop the international standard for compression of neural networks for content description and analysis (MPEG-7 part 17) ~\cite{MPEG_doc1}.

\subsection{Implementation Details}

For micro-structured weight unification, we retrained from the pre-trained full-precision model to achieve the target micro-structured block shape and unification ratio. 

As described in Subsection~\ref{sec:admm}, the regularization term is dynamically updated by our algorithm. After the training is completed, the unification threshold is applied to each block so that the absolute value of all coefficients in the selected block share the same uniform value. The micro-structured weight pruning process has a similar training procedure, where after the training is completed, the pruning threshold is applied to each block so that all coefficients in the selected block are set to zero.

Some layers may have greater impact on the prediction performance than others, such as the first feature extraction layer that usually learns detailed low-level features or the last layer that aggregates information for final prediction. Such layers may be excluded from unification or pruning for better task performance. Also, layers with negligible small amount of parameters compared with the overall model can be excluded. Also, in our experiments we used one unification ratio or pruning ratios for all chosen layers. This provided a simplified setting to clearly compare different methods. Please note that different layers can use different unification or pruning ratios, and the optimal configuration can be automatically found through methods like cross-validation, and our performance can be further improved.

For training, we used the SGD optimizer for ResNet, VGGNet and MobileNet-V2, and the Adam optimizer for Autoencoder and ConvNet. The weight decay was turned off for Autoencoder, otherwise was $1\mathrm{e}{\!-\!4}$ by default. The learning rate was $10\%$ to $50\%$ of the original settings used for the pretrained model. The batch size was 256 for ImageNet, and 128 for other datasets. 

For experiments, we first evaluate the proposed micro-structured weight unification and micro-structured weight pruning over different models and datasets with different compression ratios. Then we evaluate the performance of different block shapes and combination of unification and pruning methods. 




\subsection{Results on ImageNet}
We experimented on VGGNet-16, ResNet-50 and MobileNet-V2 over ImageNet. 
The micro-structured weight unification results are shown as ``$\bf{MWU-}\star$'', and the micro-structured weight pruning results are shown as ``$\bf{MWP-}\star$''.
For each tested model, we evaluated the performance of both micro-structured unification and micro-structured pruning using four different unification ratios or pruning ratios, to see the robustness of these algorithms with different compression ratios. 

Table~\ref{table:VGG_imagenet} gives the performance over VGGNet-16. The uncompressed pretrained model has $70.944\%$ Top-1 and $89.844\%$ Top-5 accuracy. We used $2\times2\times2\times$ blocks for convolutional layers and $2\times2$ blocks for fully connected layers. The micro-structured weight unification achieved $3.21\times$ compression ratio with $0.928\%$ Top-5 accuracy loss for VGG-16, and the micro-structured weight pruning got  $3.25\times$ compression ratio with $0.800\%$ Top-5 accuracy loss.

\begin{table}[h]
\scriptsize
\centering
\vspace{-2mm}
\caption{Results on VGGNet-16 for ImageNet dataset.}\label{table:VGG_imagenet}
\vspace{-5mm}
\begin{tabular}{p{2cm}p{1.5cm}p{1.5cm}p{1.8cm}}
\\
\toprule
Method & Top-1 Acc. & Top-5 Acc.  & Compression ratio \\ 
\midrule
Uncompressed & 70.944\%  & 89.844\%  & 1$\times$ \\ 
\bf{MWU-1}  & 69.428\%  & 88.920\%  & 2.12$\times$ \\ 
\bf{MWU-2}  & 69.314\%  & 88.820\%  & 2.54$\times$ \\ 
\bf{MWU-3}  & 69.190\%  & 88.916\%  & 3.21$\times$ \\ 
\bf{MWU-4}  & 69.338\%  & 88.920\%  & 3.75$\times$ \\ 
\bf{MWP-1}  & 69.892\%  & 89.044\%  & 3.25$\times$ \\ 
\bf{MWP-2}  & 69.276\%  & 88.910\%  & 3.88$\times$ \\ 
\bf{MWP-3}  & 68.236\%  & 88.364\%  & 4.83$\times$ \\ 
\bf{MWP-4}  & 66.818\%  & 87.218\%  & 6.43$\times$ \\ 
\bottomrule
\end{tabular} 
\vspace{-2mm}
\end{table}

Table~\ref{table:ResNet_imagenet} shows results for ResNet-50. The uncompressed pretrained baseline has  $74.970\%$ Top-1 and $92.166\%$ Top-5 accuracy. We used $2\times2\times2\times$ blocks for convolutional layers (reduced to $2\!\times\!2$ for layers with $1\!\times\!1$ kernels). We excluded the first convolutional layer and last fully-connected layer. The micro-structured weight unification achieved $3.59\times$ compression ratio with only $0.962\%$ Top-5 accuracy loss, while the micro-structured weight pruning got $3.45\times$ compression ratio with $2.572\%$ Top-5 accuracy loss. 

\begin{table}[h]
\scriptsize
\centering
\vspace{-2mm}
\caption{Results on ResNet-50 for ImageNet dataset.}\label{table:ResNet_imagenet}
\vspace{-5mm}
\begin{tabular}{p{2cm}p{1.5cm}p{1.5cm}p{1.8cm}}
\\
\toprule
Method & Top-1 Acc. & Top-5 Acc.  & Compression ratio \\ 
\midrule
Uncompressed & 74.970\%  & 92.166\%  & 1$\times$ \\ 
\bf{MWU-1}  & 74.148\%  & 91.692\%  & 2.09$\times$ \\ 
\bf{MWU-2}  & 73.836\%  & 91.536\%  & 2.42$\times$ \\ 
\bf{MWU-3}  & 73.636\%  & 91.536\%  & 2.99$\times$ \\ 
\bf{MWU-4}  & 73.100\%  & 91.204\%  & 3.59$\times$ \\ 
\bf{MWP-1}  & 73.662\%  & 91.710\%  & 2.36$\times$ \\ 
\bf{MWP-2}  & 73.072\%  & 91.274\%  & 2.64$\times$ \\ 
\bf{MWP-3}  & 71.862\%  & 90.638\%  & 2.99$\times$ \\ 
\bf{MWP-4}  & 70.200\%  & 89.594\%  & 3.45$\times$ \\ 
\bottomrule
\end{tabular}
\vspace{-2mm}
\end{table}

Table~\ref{table:MobileNet_imagenet} shows results for MobileNet-V2. It is challenging to prune MobileNet-V2 due to the compact architecture specially designed for mobile devices. We excluded the first layer and all depth-wise convolutional layers as the total parameters of all depth-wise convolutional layers accounted for less than 2\% of the total weights. We used $2\times2$ blocks for both convolutional layers and fully-connection layer. Results in Table~\ref{table:MobileNet_imagenet} show that  our micro-structured weight unification method achieved $2.41\times$ compression ratio with $2.966\%$ Top-5 accuracy loss. In similar compression settings, micro-structured weight pruning got $2.23\times$ compression ratio with $3.386\%$ Top-5 accuracy loss.

\begin{table}[h]
\scriptsize
\centering
\caption{Results on MobileNet-V2 for ImageNet dataset.}\label{table:MobileNet_imagenet}
\vspace{-5mm}
\begin{tabular}{p{2cm}p{1.5cm}p{1.5cm}p{1.8cm}}
\\
\toprule
Method & Top-1 Acc. & Top-5 Acc.  & Compression ratio \\ 
\midrule
Uncompressed & 71.488\%  & 90.272\%  & 1$\times$ \\ 
\bf{MWU-1}  & 67.110\%  & 87.918\%  & 2.02$\times$ \\ 
\bf{MWU-2}  & 66.272\%  & 87.306\%  & 2.41$\times$ \\ 
\bf{MWU-3}  & 65.248\%  & 86.836\%  & 2.96$\times$ \\ 
\bf{MWU-4}  & 63.336\%  & 85.662\%  & 3.62$\times$ \\ 
\bf{MWP-1}  & 70.634\%  & 89.898\%  & 1.36$\times$ \\ 
\bf{MWP-2}  & 69.432\%  & 89.444\%  & 1.56$\times$ \\ 
\bf{MWP-3}  & 67.534\%  & 88.192\%  & 1.83$\times$ \\ 
\bf{MWP-4}  & 65.416\%  & 86.886\%  & 2.23$\times$ \\ 
\bottomrule
\end{tabular}
\vspace{-2mm}
\end{table}

In summary, micro-structured unification and micro-structured pruning perform similarly on VGGNet-16, which has a lot of redundancy in the network architecture. For more compact networks like ResNet-50 and MobileNet-V2, the micro-structured weight unification shows clear advantages over the micro-structured weight pruning. The observations confirm that micro-structured unification can provide a compressed model with balanced compression ratio and task performance, and can work robustly for networks that are redundant or compact in design. 

\subsection{Results on CIFAR-100 and DCASE}

We conducted experiment on Autoencoder using the CIFAR-100 dataset and ConvNet using the DCASE dataset, respectively. Both models are quite small in size (less than 500KB), one for image compression and one for audio classification. The datasets and models are used by MPEG to evaluate the generalization ability of different network compression methods over different tasks. The Autoencoder is measured by PSNR and SSIM, which are the most widely used metrics for image quality evaluation. The DCASE audio classification is measured by Top-1 accuracy. 

Table~\ref{table:Autoencoder_DCASE} shows results for the image compression and audio classification tasks. We also tested four different unification or pruning ratios, and used $2\times2$ blocks for convolutional layers. From the table, for image compression, our micro-structured weight unification even improved the performance of the baseline model while compressing the model at the same time (with a $1.31\times$ compression ratio). For audio classification, our micro-structured weight unification achieved a $5.70\times$ compression ratio and at the same time improved the performance of the baseline model too (with a $60.247\%$ Top-1 accuracy).

\begin{table}[h]
\scriptsize
\centering
\vspace{-2mm}
\caption{Results on Autoencoder for CIFAR-100 dataset and ConvNet for DCASE dataset.}\label{table:Autoencoder_DCASE}
\vspace{-2mm}
\resizebox{\columnwidth}{!}{%
\begin{tabular}{c|ccc|cc}

\toprule
 & \multicolumn{3}{c|}{Autoencoder} & \multicolumn{2}{c}{ConvNet} \\
\toprule
 & & & Compression & & Compression \\ 
Method & PSNR  & SSIM  & ratio & Top-1 Acc.  & ratio \\ 
\midrule
Uncompressed & 30.134 &  0.956 & 1$\times$ & 58.272\%  & 1$\times$ \\ 
\bf{MWU-1}  & 30.398 & 0.958 & 1.31$\times$  & 61.148\%  & 2.48$\times$\\ 
\bf{MWU-2}  & 29.845 & 0.952 & 1.66$\times$  & 61.111\%  & 3.13$\times$\\ 
\bf{MWU-3}  & 28.475 & 0.937 & 4.16$\times$  & 60.741\%  & 4.29$\times$\\ 
\bf{MWU-4}  & 28.213 & 0.933 & 5.75$\times$  & 60.247\%  & 5.70$\times$\\ 
\bf{MWP-1}  & 29.549 & 0.951 & 1.45$\times$  & 61.111\%  & 2.86$\times$ \\ 
\bf{MWP-2}  & 29.324 & 0.949 & 2.22$\times$  & 60.864\%  & 4.03$\times$\\ 
\bf{MWP-3}  & 29.028 & 0.946 & 5.04$\times$  & 59.630\%  & 5.08$\times$\\ 
\bf{MWP-4}  & 28.489 & 0.941 & 6.84$\times$ & 58.519\%  & 6.04$\times$ \\ %
\bottomrule
\end{tabular}
}
\vspace{-2mm}
\end{table}



\subsection{Influence of Block Shapes}
\label{experiment:blockshape}

\begin{figure}[h]
    \centering
    \includegraphics[width=0.4 \textwidth]{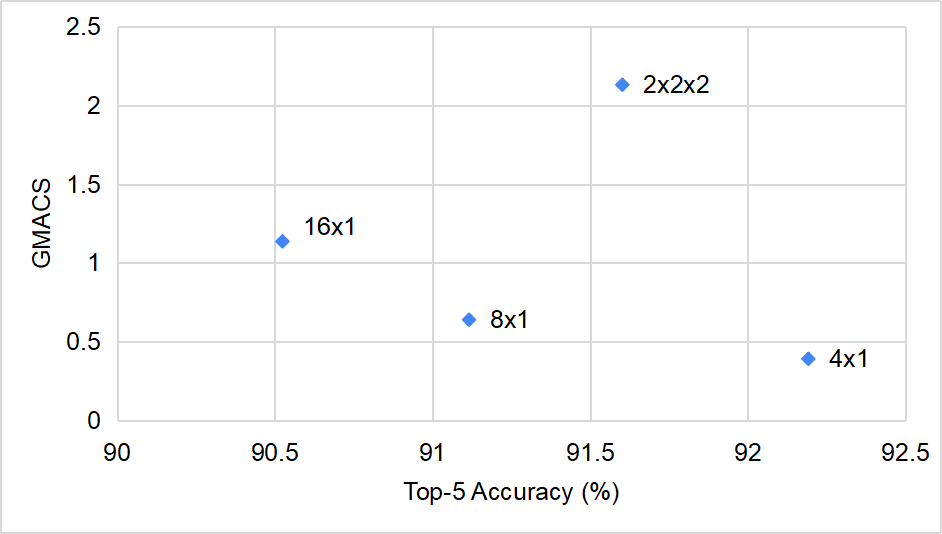}
    \vspace{-1mm}
    \caption{Comparisons of different block shapes for micro-structured weight Unification}
    \label{fig:diff_shape_MWU}
    \vspace{-2mm}
\end{figure}

As mentioned in the Section~\ref{sec:swu} and Table~\ref{tab:patternshape}, the unification block shape is one key factor affecting the overall model reduction and computation acceleration. Since the Res-block is one of the most widely used module in modern DNNs, we chose ResNet-50 to test the effect of using different block shapes. Figure~\ref{fig:diff_shape_MWU} shows the performance and complexity comparison of different block shapes.

The original ResNet-50 has $4.11$ GMACS for overall  computation (with $224\!\times\!224$ inputs). We tested four different block shapes, $2\times2\times2$, $4\times1$, $8\times1$ and $16\times1$. For the $2\times2\times2$ configuration, we used $2\times2$ block shape for convolutional layers with $1\times1$ kernels, and $2\times2\times2$ block shape for convolutional layers with $3\times3$ kernels. 
The first layer and last layer were excluded from the training process. 
From Figure~\ref{fig:diff_shape_MWU}, it is easy to see that better accuracy was achieved by using the 3-dim block shape. However, the estimated computation was $2.12$ GMACS for $2\times2\times2$ blocks, which was much higher than the $0.64$ GMACS for $8\times1$ blocks. On the other hand, the $8\times1$ block shape achieved better compression ratio ($4.76\times$) than that of $2\times2\times2$ blocks ($3.59\times$), with similar accuracy too (only $0.486\%$ and $0.590\%$ difference in Top-5 and Top-1 accuracy). 
In other words, the 2-dim block shape can be more preferable if our goal is to reduce computation, as it is in general more difficult to train a model with higher block dimensions. 

The results also show that the performance decrease sharply as the block size increases, which is consistent with our previous analysis. The larger the blocks, the stronger constrains we put on unified weights, the harder it is to maintain the original performance.

\subsection{Combination Test}

As shown in the above experiments, micro-structured weight unification and weight pruning have different strengths for different situations. In general, weight unification aims for a balance between the original task performance and reduced computation, while weight pruning is the aggressive special case of weight unification pursuing compression effects more. 
Actually, these two methods are complementary to each other and can be combined to achieve a better overall performance. 

In this subsection, we experimented on combining the micro-structured weight unification and weight pruning. Again, we tested the combined method on ResNet-50 using the ImageNet dataset. We used a simple setting for evaluation here, i.e., applying micro-structured weight unification or weight pruning in a layer-wise manner. Specifically, $8\times1$ blocks with $100\%$ unification ratio were used for micro-structured weight unification and $1\times4$ blocks with $50\%$ prune ratio in each block were used for weight pruning. This pruning ratio is recommended as the best setting for Nvidia A100 Tensor Core sparsity acceleration~\cite{nvidai_a100}. 
These block shapes were empirically selected for their overall robust performance across different models and tasks.


We applied micro-structured weight pruning to the first and last layers, and the first and last convolutional layers with kernel size $1\times1$ in each residual block. The micro-structured weight unification was applied to all the remaining layers. This compression setting was empirically determined based on cross-validation. Finally, we achieved $3.02\times$ compression ratio with $1.020\%$ Top-1 accuracy loss and $0.436\%$ Top-5 accuracy loss compared to the uncompressed pretrained baseline. The promising results demonstrate that combining different compression methods can make use of their different strengths and potentially bring further performance improvements.

\section{Conclusion}
In this paper, we proposed a hardware-friendly micro-structured weight unification framework to achieve large amount of storage reduction and inference acceleration, while maintaining the original task performance at the same time. We incorporated the ADMM algorithm to effectively train our compressed model by relaxing the hard micro-structured weight constraints and iteratively solving the decomposed dynamic regularization terms. The general micro-structured weight unification mechanism and the corresponding training framework can be flexibly applied to various network architectures for different tasks. Experimental results over several benchmarks of different tasks showed that our method can achieve significant compression ratio with greatly reduced computation.

The micro-structured weight unification and its aggressive special version, micro-structured weight pruning, can be further combined to exploit the advantages of both methods. Also, different micro-structured block shapes can be used for different models and layers and for different computing engines. We use simplified settings in this paper to clearly compare different methods, which does not fully exert our potential. The optimal configurations, i.e., block shapes, unification or pruning methods, unification or pruning ratios, etc., vary according to different models and tasks. With better configurations, automatically determined (e.g., through cross-validation) or empirically designed, further performance improvements can be expected.

{\small
\bibliographystyle{ieee_fullname}
\bibliography{egbib}
}

\end{document}